%% file: arxiv.tex
\documentclass{article}

 \usepackage[preprint,nonatbib]{neurips_2025}

\usepackage[utf8]{inputenc} 
\usepackage[T1]{fontenc}    
\usepackage{hyperref}       
\usepackage{url}            
\usepackage{booktabs}       
\usepackage{amsfonts}       
\usepackage{nicefrac}       
\usepackage{microtype}      

\usepackage{amsmath}
\usepackage[utf8]{inputenc} 
\usepackage[T1]{fontenc}    
\usepackage{hyperref}       
\usepackage{url}            
\usepackage{booktabs}       
\usepackage{amsfonts}       
\usepackage{nicefrac}       
\usepackage{microtype}      
\usepackage[dvipsnames]{xcolor}         
\usepackage{wrapfig}
\usepackage{caption}
\usepackage{amsmath}
\usepackage{subcaption}
\usepackage{bbding}
\usepackage{graphicx}
\usepackage{multirow}
\usepackage{tabu}

\usepackage{tcolorbox}
\usepackage{colortbl}
\usepackage{geometry}
\tcbuselibrary{breakable}
\usepackage{listings}

\newcommand{\ie}{\textit{i}.\textit{e}.}
\newcommand{\eg}{\textit{e}.\textit{g}.}

\AtEndPreamble{
    \usepackage[capitalize]{cleveref}
    \crefname{section}{Sec.}{Secs.}
    \Crefname{section}{Section}{Sections}
    \Crefname{table}{Table}{Tables}
    \crefname{table}{Tab.}{Tabs.}
}
\usepackage{colortbl}
\definecolor{mygray}{gray}{0.92}
\definecolor{my_green}{RGB}{82,208,80}
\usepackage{enumitem}
\usepackage{makecell}
\definecolor{00red}{RGB}{236,35,35}

\usepackage{arydshln}

\usepackage{pifont}

\title{
R1-ShareVL: Incentivizing Reasoning Capability of Multimodal Large Language Models via Share-GRPO
}

\author{%
  Huanjin Yao$^{2,3}\textsuperscript{*}$, 
  Qixiang Yin$^{4}\textsuperscript{*}$, 
  Jingyi Zhang$^1$,
  Min Yang$^2$,
  Yibo Wang$^3$,
  Wenhao Wu$^{5}$ \\
  \textbf{Fei Su$^4$,
  Li Shen$^1$,
  Minghui Qiu$^2$,
  Dacheng Tao$^{1}$,
  Jiaxing Huang$^{1}\textsuperscript{\Envelope}$ } \\
  $^1$Nanyang Technological University \quad 
  $^2$ByteDance \quad
  $^3$Tsinghua University \quad \\
  $^4$Beijing University of Posts and Telecommunications \quad
  $^5$The University of Sydney 
  \\ 
  {\small $^{*}$ Equal Contribution \qquad \Envelope~Corresponding Author} \\
}

\begin{document}

\maketitle

\begin{abstract}
In this work, we aim to incentivize the reasoning ability of Multimodal Large Language Models (MLLMs) via reinforcement learning (RL) and develop an effective approach that mitigates the sparse reward and advantage vanishing issues during RL. To this end, we propose Share-GRPO, a novel RL approach that tackle these issues by exploring and sharing diverse reasoning trajectories over expanded question space. Specifically, Share-GRPO first expands the question space for a given question via data transformation techniques, and then encourages MLLM to effectively explore diverse reasoning trajectories over the expanded question space and shares the discovered reasoning trajectories across the expanded questions during RL. In addition, Share-GRPO also shares reward information during advantage computation, which estimates solution advantages hierarchically across and within question variants, allowing more accurate estimation of relative advantages and improving the stability of policy training. Extensive evaluations over six widely-used reasoning benchmarks showcase the superior performance of our method. Code will be available at \url{https://github.com/HJYao00/R1-ShareVL}.
\end{abstract}

\section{Introduction}

The recent success of Reinforcement Learning (RL) in Large Language Models (LLMs), such as Kimi-K1.5~\cite{team2025kimi} and DeepSeek-R1~\cite{guo2025deepseek}, shows its promise in incentivizing model's long-chain reasoning capability, enabling LLMs to tackle complex tasks such as mathematical and scientific reasoning. 
The core design of these advances (\eg, GRPO~\cite{shao2024deepseekmath} in Deepseek-R1) lies in online reinforcement learning without the need of reward models, which encourages an LLM to generate a group of reasoning paths and iteratively refine its reasoning process with a group relative advantage estimation mechanism based on rule-based reward functions. 
Typically, a simple reward strategy is adopted: reasoning paths leading to correct answers receive higher rewards, while those leading to incorrect answers receive lower ones, where the model is optimized via the group relative advantages estimated from the rewards.

Inspired by these advancements, we aim to develop a simple and effective reinforcement learning method for Multimodal LLMs (MLLMs) to incentivize their long-chain reasoning ability.
A simple way is to directly apply these LLM online reinforcement learning methods like GRPO on MLLMs. 
However, we empirically observe that directly applying GRPO on MLLMs suffers from sparse reward and advantage vanishing issues, leading to degraded performance in enhancing MLLM’s reasoning capability~\cite{r1-vl,meng2025mm,wang2025vl}:

\textbf{{(1) Sparse reward:}} Most current MLLMs, especially smaller ones, exhibit very limited long-chain reasoning capability. As a result, only a few generated reasoning paths receive positive rewards, especially on challenging questions and particularly during the early stage of training. This leads to sparse rewarding, inefficient exploration and instable training in GRPO-like methods.

\textbf{{(2) Advantage vanishing:}} GRPO-like methods compute relative advantages by comparing the rewards of a group of responses sampled from a given question, leading to advantage vanishing when receiving homogeneous responses.
Specifically, along reinforcement learning process, the model tends to gradually predict similar and all correct responses for well-learned questions, and similar and all incorrect responses for poor-learned questions. 
In this way, the relative advantages tend to approach zero when the group of responses become more homogeneous, and collapse to zero when all responses receive identical rewards (\eg, all correct or all incorrect), resulting ineffective reinforcement learning.

\begin{figure}[!t]
  \centering
  \includegraphics[width=1\linewidth]{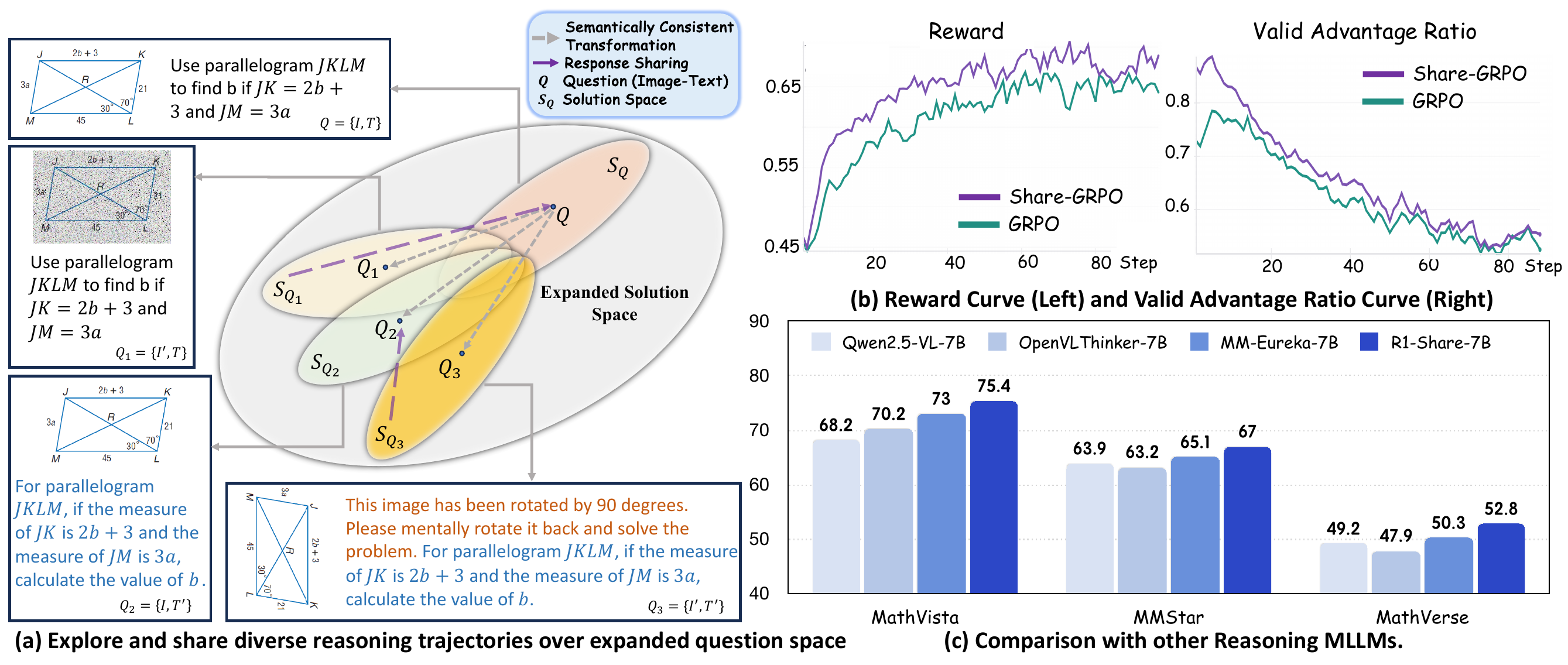}
  \caption{
\textbf{(a)} Share-GRPO expands the question space via semantically consistent transformations, and then explores diverse reasoning trajectories from different question variants and shares the discovered trajectories among them.
\textbf{(b)} Share-GRPO provides denser rewards and higher valid advantage ratios compared to GRPO, demonstrating its effectiveness in mitigating sparse reward and advantage vanishing issues.
\textbf{(c)} Share-GRPO outperforms the baseline and other SOTA RL-based reasoning MLLMs on both mathematical and general reasoning benchmarks.
}
\vspace{-2mm}
  \label{fig: fig1}
\end{figure}

Motivated by these observations, we propose Share-GRPO, a novel approach that introduces the concept of information sharing into MLLM reinforcement learning to mitigate sparse reward and advantage vanishing issues.
The core idea of Share-GRPO lies in exploring and sharing diverse reasoning trajectories over expanded question space as shown in Fig.~\ref{fig: fig1} (a). 
Specifically, Share-GRPO first expands the question space for a given question via data transformation techniques, and then encourages MLLM to effectively explore diverse reasoning trajectories over the expanded question space and shares the discovered reasoning trajectories across the expanded questions during reinforcement learning.
In this way, each expanded question variant can both contribute and benefit from the reasoning trajectories generated by others in the expand question space, allowing the model to jointly explore and learn from a shared solution space across expanded questions.

In addition, Share-GRPO also shares reward information during advantage computation, which estimates solution advantages hierarchically across and within question variant, allowing more accurate estimation of relative advantages and improving the stability of policy training.
Specifically, we estimate advantages at two levels: a local level, which consists of responses generated from each individual question variant, and a global level, which aggregates responses across all variants of the same seed question.
This hierarchical advantage estimation enables more robust and fine-grained relative advantage computation, where the local level captures intra-variant structure and variance while the global level exploits cross-variant diversity and complementarity and stabilizes reward signals.

In this way, Share-GRPO effectively mitigates the sparse reward and advantage vanishing issues: 
(1) Share-GRPO expands the question space and enables more diverse solution space for each given question, which effectively increases the likelihood of generating a successful reasoning response and thus mitigates the sparse rewarding issue as illustrated in the left curve of Fig.~\ref{fig: fig1} (b).
(2) Share-GRPO allows the model to explore diverse reasoning trajectories from the expanded question space and shares the discovered reasoning trajectories, ultimately mitigating the advantage vanishing issue effectively as illustrated in the  right curve in Fig.~\ref{fig: fig1} (b).
(3) Share-GRPO estimates solution advantages hierarchically across and within question variant, which enables more accurate estimation of relative advantages and stable reinforcement learning process.

In summary, the main contributions of this work are summarized as follows: 
First, we introduce the concept of information sharing into MLLM reinforcement learning, and propose Share-GRPO which explores and shares diverse reasoning trajectories over expanded question space, effectively mitigating the sparse reward and advantage vanishing issues.
To the best of our knowledge, this is the first work that explores information sharing for MLLM reasoning reinforcement learning.
Second, we design a hierarchical advantage estimation method by sharing reward information, which estimates solution advantages hierarchically across and within question variant, allowing accurate and robust advantage estimation.
Third, extensive experiments on 6 MLLM reasoning benchmarks demonstrate the superiority of our proposed methods as illustrated in Fig.~\ref{fig: fig1} (c).

\section{Related Work}

\label{gen_inst}

\subsection{Multimodal Large Language Model}

Multimodal Large Language Models (MLLMs) ~\cite{achiam2023gpt,bai2023qwen,yang2024qwen2,tong2024cambrian,chen2024internvl,lu2025internvl,team2023gemini,yao2024minicpm,yao2024dense,valley2} demonstrate outstanding performance in semantic understanding of cross-domain visual content and multimodal reasoning.
Early research on MLLMs primarily focused on text-image alignment and the integration of multiple modalities ~\cite{liu2023visual,alayrac2022flamingo,liu2024improved,lan2025gem,liu2024llavanext}. Subsequently, models like GPT-4V~\cite{yang2023dawn} achieved breakthroughs in cross-modal understanding through multimodal instruction fine-tuning, enabling them to support simple tasks such as image captioning, visual question answering and OCR. More complex tasks, such as mathematical reasoning, document understanding, etc., require MLLMs to be able to perform complex logical deductions. For MLLM reasoning, models such as Multimodal-CoT~\cite{zhang2023multimodal} and LLaVA-CoT~\cite{xu2411llava} employ chain-of-thought (CoT) reasoning, breaking down the multimodal reasoning process into step-by-step inference steps while leveraging multimodal data to improve the model’s reasoning capabilities.
Additionally, Mulberry~\cite{yao2024mulberry} proposes CoMCTS to generate effective reasoning paths through multi-model collaboration.
Different from these studies, this work focuses on reinforcement learning to improve MLLM reasoning capability.

\subsection{Reinforcement Learning for Multimodal Large Language Model Reasoning}

Reinforcement learning has become an essential technology for enhancing the capabilities of MLLMs. Early research primarily focused on Reinforcement Learning from Human Feedback(RLHF)~\cite{luo2025o1,lee2023rlaif,yu2024rlhf,yu2024rlaif}, which aligns the outputs of multimodal models with human preferences by incorporating human feedback signals. Recently, DeepSeek-R1~\cite{guo2025deepseek} utilizes a simple rule-based reward function to provide effective and reliable reward signals during the RL process. This indicates that the Group Relative Policy Optimization (GRPO) with result-level rewards effectively enhances the reasoning ability of LLMs. In the multimodal domain, researchers have begun exploring the use of RL to enhance the visual reasoning capabilities of MLLMs.  Recent works, such as Vision-R1~\cite{huang2025vision} and MM-Eureka~\cite{meng2025mm} have open-sourced large-scale SFT cold start data and RL data. R1-V~\cite{chen2025r1v}, Reason-RFT~\cite{tan2025reason}, R1-VL~\cite{r1-vl} and other methods~\cite{yang2025r1,chen2025sft,deng2025openvlthinker,peng2025skywork, vision-r1-zhan} have designed various rule-based reward functions to enhance the reasoning abilities of MLLMs, such as geometric understanding and spatial perception. 
Unlike these methods, our ShareGRPO explores information sharing for MLLM reasoning reinforcement learning to mitigate sparse reward and advantage vanishing issues.

\subsection{Information Sharing in Deep Learning}

Information sharing is a key strategy in deep learning, enabling more effective learning through the exchange of signals across modalities, tasks, or hierarchical model components.
In multi-modal learning, models such as ViLBERT~\cite{lu2019vilbert} and LXMERT~\cite{tan2019lxmert} employ cross-modal attention to achieve fine-grained information fusion between vision and language streams.
In contrastive learning (\eg, SimCLR~\cite{simclr}, MoCo~\cite{moco}), shared representations across augmented views enhance feature robustness.
This concept extends to reinforcement learning, especially in multi-task and multi-agent settings, where information sharing improves sample efficiency and mitigates sparse rewards. Methods like Distral~\cite{teh2017distral} and PopArt~\cite{popart} promote shared policy structures, while agents in multi-agent RL benefit from shared value functions or communication protocols~\cite{foerster2016learning, lowe2017multi}. \cite{d2024sharing} further demonstrate that shared representations enhance generalization in multi-task RL.
Unlike prior work, we introduce information sharing into MLLM reasoning reinforcement learning to mitigate sparse rewards and advantage vanishing for more effective reasoning learning.

\section{Method} 
\label{sec: method} 

This section first provides the preliminary of Group Relative Policy Optimization (GRPO), and then presents the proposed Share-GRPO that introduces the concept of information sharing into MLLM reinforcement learning. Further details are elaborated in the subsequent subsections.

\subsection{Preliminary} 

\textbf{Group Relative Policy Optimization (GRPO)}. GRPO~\cite{shao2024deepseekmath} is a variant of Proximal Policy Optimization (PPO)~\cite{schulman2017proximal}, designed to enhance the performance of LLMs on complex reasoning tasks, such as mathematical and scientific reasoning. 
Starting with a pretrained MLLM to be optimized, GRPO first uses it to initialize a policy model $\pi_{\theta}$ and a reference model $\pi_{\text{old}}$. For a given image-text pair $(I, T)$, the reference policy model $\pi_{\theta_{\mathrm{old}}}$ generates a set of responses $\{{o}_1, {o}_2, ..., {o}_G\}$. A group-based reward function then computes the corresponding rewards $\{R_1, R_2, ..., R_G\}$, which are subsequently used to estimate the advantage $\hat{A}{i}$ for each response relative to the group: 

\vskip -0.1in
\begin{align}
\hat{A}_{i} = \frac{R_i -\text{mean}\left(\{R_i\}_{i=1}^{G}\right)}{\text{std}\left(\{R_i\}_{i=1}^{G}\right)}.
\end{align}

Similar to PPO, GRPO employs a clipped objective with a KL penalty term: 
\begin{align}
 &\mathcal{J}_{\mathrm{GRPO}}(\theta) = {} \mathbb{E}_{(I,T)\sim p_{\mathcal{D}},{o}\sim\pi_{\theta_\text{old}}(\cdot|I,T)}\nonumber\\
&\Biggl[ \frac{1}{n}\sum_{i=1}^{n} \min\ \!\Biggl(\frac{\pi_{\theta}({o}_i \mid I,T)}{\pi_{\theta_{\mathrm{old}}}({o}_i \mid I,T)}\hat{A}_i, \mathrm{clip}\ \!\Bigl(\frac{\pi_{\theta}({o}_i \mid I,T)}{\pi_{\theta_{\mathrm{old}}}({o}_i \mid I,T)},\,1-\epsilon,\,1+\epsilon\Bigr)\hat{A}_i - \beta D_{\mathrm{KL}}\left(\pi_{\theta}| | \pi_{\mathrm{ref}}\right)\Biggr) \Biggr]\textrm{.}
\end{align}

\textbf{Sparse Reward and Advantage Vanishing Issues.} 
Despite the effectiveness of GRPO,  it generally faces two challenges when applied to MLLMs: the sparse reward issue and the advantage vanishing issue. 
Sparse rewarding arises due to the limited reasoning ability of current MLLMs, where only a few reasoning paths receive positive rewards, leading to inefficient exploration and instable training. To alleviate this, prior work such as R1-VL~\cite{r1-vl} introduces step-wise reward signals to provide dense rewards throughout the reasoning process.
Advantage vanishing occurs when MLLMs generate homogeneous responses for the same question and receive identical rewards, causing the relative advantages to collapse to zero and resulting in ineffective reinforcement learning. 
To tackle this issue, VL-Rethinker~\cite{wang2025vl} and Skywork R1~\cite{wei2025skywork} select the samples with large magnitudes of advantages and reuse them in RL process, while MM-Eureka~\cite{meng2025mm} employs an online filtering strategy to remove the samples with zero advantage. 
Different from the prior works, our Share-GRPO effectively addresses both of these two challenges by exploring and sharing diverse reasoning trajectories over expanded question space, therefore encouraging reward diversity and stable policy optimization.

\begin{figure}[!t]
  \centering
  \includegraphics[width=1\linewidth]{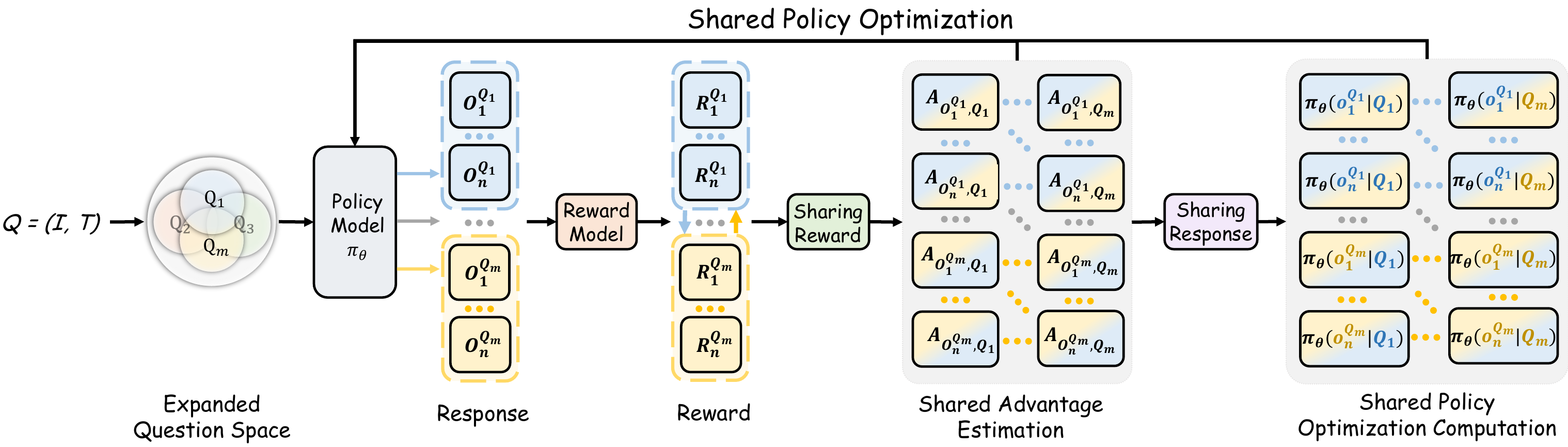}
  \vspace{-2mm}
  \caption{
  Overview of the proposed Share-GRPO.
  For a given question, Share-GRPO first applies semantically consistent transformation to generate a set of varied but semantically equivalent questions, thereby expanding the question space. It then encourages the MLLM to explore diverse reasoning paths over the expanded question space and facilitates the sharing of discovered reasoning trajectories and their rewards across the expanded questions during the reinforcement learning process.
  }
  \label{fig: fig2}
  \vskip -0.1in
\end{figure}

\subsection{Share-GRPO}

We propose Share-GRPO, a novel online MLLM reinforcement learning framework that mitigates the sparse reward and advantage vanishing issues via exploring and sharing diverse reasoning trajectories over expanded question space.
Specifically, for a given question, Share-GRPO first applies semantically consistent transformation to generate a set of varied but semantically equivalent questions, thereby expanding the question space. It then encourages the MLLM to explore diverse reasoning paths over the expanded question space and facilitates the sharing of discovered reasoning trajectories and their rewards across the expanded questions during the reinforcement learning process, as illustrated in Fig.~\ref{fig: fig2}.

\subsubsection{Reasoning Space Expansion}

\textbf{Question Space Expansion.}
To expand the question space for a given question, we introduce Semantically Consistent Transformation (SCT) which generates a group of question variant $\mathbf{Q} = \{Q_1, Q_2, ..., Q_m\}$ for each given question $Q_{ori}=\{T_{ori}, I_{ori}\}$.
Specifically, we propose two types of transformation techniques, \ie, offline textual SCT and online multimodal SCT, for more diverse, comprehensive and flexible question space expansion.

\textit{(1) Offline Textual Semantically Consistent Transformation.}
Prior to online reinforce learning, we first employ offline textual SCT $\phi(\cdot)$ to rewrite the textual prompt  $T_{ori}$ for each give question.
Specifically, we prompt GPT-4o to generate $m$ semantically consistent variants, resulting in an expanded question set. 
The textual prompts of the generated variants differ from that of the original question $T_{ori}$ in syntactic structure and lexical expressions, while preserving the original intent and the corresponding correct answer:

\vskip -0.1in
\begin{align}
Q^{\text{offline}} = \{\phi(T_{ori}), I_{ori}\}.
\end{align}

\textit{(2) Online Multimodal Semantically Consistent Transformation.}
During online reinforcement learning, we introduce a multimodal SCT strategy to further expand the question space on the fly.
Given an image $I_{ori}$ in the input question, we apply visual transformations $\psi(\cdot)$ to alter its visual content.
Specifically, we carefully select transformations (\eg, rotation, noise injection) that preserve critical visual cues necessary for reasoning, and avoid transformations (\eg, cropping, color distortion) that may disrupt key information.
Each image undergoes one randomly selected transformation with a probability $p$.
 
In addition, to mitigate the potential semantic inconsistencies between the visual and textual inputs after visual changes, we perform a manual textual transformation $\tau$ that appends a transformation-specific prompt to the corresponding textual prompt, providing contextual guidance aligned with the visual modification:

\begin{equation}
Q^{\text{online}} = \{\tau(\phi(T_{ori})), \psi(I_{ori})\}.
\end{equation}

\textbf{Solution Space Expansion.}
With the expanded question space $\mathbf{Q} = \{Q_1, Q_2, ..., Q_m\}$, Share-GRPO enables to explore diverse reasoning trajectories in an enlarged solution space for each given question. 
Specifically, for each question $Q_i \in \mathbf{Q}$, the policy model $\pi_{\theta}$ generates $n$ candidate reasoning responses, resulting in an expanded response set: $\mathbf{O}=\{\{o_{1}^{Q_1}, ..., o_{n}^{Q_1}\}, ..., \{o_{1}^{Q_m}, ... o_{n}^{Q_m}\}\}$.

\subsubsection{Shared Advantage Estimation}

With the expanded reasoning space, Share-GRPO shares reward information during advantage computation, which estimates reasoning trajectory advantages hierarchically across and within question variant. 

Following GRPO~\cite{shao2024deepseekmath}, we adopt rule-based reward functions to compute the reward for each generated reasoning trajectory, \ie, $R=\{\{r_{1}^{Q_1}, ..., r_{n}^{Q_1}\}, ..., \{r_{1}^{Q_m}, ... r_{n}^{Q_m}\}\}$.
Specifically, we adopt an outcome-level accuracy reward, which assigns higher rewards to reasoning paths that lead to correct answers and lower rewards to those leading to incorrect ones. In addition, we employ a format reward that encourages the reasoning trajectory to follow a detailed step-by-step process before providing the final answer.

With the computed rewards $R$, we propose a hierarchical advantage estimation approach that computes advantage at two levels: a global level, which aggregates responses across all variants of the same original question; and a local level, which considers responses generated from each individual question variant.

\textit{(1) Global-level Advantage Estimation.} We first estimate the advantage from a global perspective, where the relative advantage is computed using the rewards obtained from all question variants $\mathbf{Q} = \{Q_1, Q_2, ..., Q_m\}$:

\vskip -0.1in
\begin{align}
\hat{A}_{i, j, k}^{\text{global}} = \frac{R_i^{Q_j} -\text{mean}\left(\{\{r_{1}^{Q_1}, ..., r_{n}^{Q_1}\}, ..., \{r_{1}^{Q_m}, ... r_{n}^{Q_m}\}\}\right)}{\text{std}\left(\{\{r_{1}^{Q_1}, ..., r_{n}^{Q_1}\}, ..., \{r_{1}^{Q_m}, ... r_{n}^{Q_m}\}\}\right)}.
\label{eq5}
\end{align}

\textit{(2) Local-level Advantage Estimation.} We also estimate the advantage at a local level, where the relative advantage is computed within the responses generated from each individual question variant $Q_j \in \mathbf{Q}$. Specifically, for each question variant $Q_j$, the local advantage is estimated as follows:

\vskip -0.1in
\begin{align}
\hat{A}_{i, j, k}^{\text{local}} = \frac{R_i^{Q_j} -\text{mean}\left(\{r_{1}^{Q_j}, ... r_{n}^{Q_j}\}\right)}{\text{std}\left(\{r_{1}^{Q_j}, ... r_{n}^{Q_j}\}\right)}.
\label{eq6}
\end{align}

With the global-level advantage and local-level advantage estimated via Eqs.~\ref{eq5} and \ref{eq6}, we can obtain the final advantage as follow:

\begin{equation}
\hat{A}_{i, j, k}^{\text{hier}} =
\begin{cases} 
    \hat{A}_{i, j, k}^{\text{global}} + \hat{A}_{i, j, k}^{\text{local}}, & j=k, \\
    \hat{A}_{i, j, k}^{\text{global}}, & j\neq k, 
\end{cases}
\label{eq7}
\end{equation}
where the local advantage $\hat{A}_{i, j, k}^{\text{local}}$ is only computed when the responses are generated from the same question variant, \ie, when $j=k$.
By incorporating hierarchical advantage estimation, Share-GRPO achieves more accurate relative advantage computation, leading to more stable and effective policy training.

\subsubsection{Shared Policy Optimization}

With the expanded reasoning space and the shared advantage estimation, Share-GRPO enables to explore and share diverse reasoning trajectories and allows more accurate advantage estimation for each given question.
Then, we optimize policy model $\pi_{\theta}$ by sharing diverse reasoning trajectories $\mathbf{O}=\{\{o_{1}^{Q_1}, ..., o_{n}^{Q_1}\}, ..., \{o_{1}^{Q_m}, ... o_{n}^{Q_m}\}\}$ across question variants $\mathbf{Q} = \{Q_1, Q_2, ..., Q_m\}$:

\vskip -0.1in
\begin{align}
L_{\mathrm{}}&(\theta) = {} \mathbb{E}_{(Q)\sim p_{\mathcal{D}},{o}\sim\pi_{\theta_\text{old}}(\cdot|Q)}\nonumber\\
&\Biggl[ \frac{1}{n}\sum_{i=1}^{n} \frac{1}{m^2}\sum_{\substack{j=1 \\ k=1}}^{m} \min\ \!\Biggl
(\frac{\pi_{\theta}({o_i^{Q_j}}  \mid {Q_k})}{\pi_{\theta_{\mathrm{old}}}({o_i^{Q_j}} \mid {Q_k})} \hat{A}_{i, j, k}^{\text{hier}},
\mathrm{clip}\ \!\Bigl(\frac{\pi_{\theta}({o_i^{Q_j}} \mid {Q_k})}{\pi_{\theta_{\mathrm{old}}}({o_i^{Q_j}} \mid {Q_k})},\,1-\epsilon,\,1+\epsilon\Bigr)\hat{A}_{i, j, k}^{\text{hier}} \Biggr) \Biggr].
\end{align}
\vskip -0.2in

\section{Experiments}
\setlength{\tabcolsep}{3pt}

In this section, we first provide implementation details in~\cref{subsec:Implementation Details}, and then present main results in \cref{sec:exp_result} that demonstrate the effectiveness of Share-GRPO.
In \cref{sec:exp_ablation}, we conduct comprehensive ablation studies to examine the impact of each design in Share-GRPO.
\cref{sec:exp_discussion} provides more discussion and analysis of Share-GRPO.
More details are elaborated in the subsequent subsections.

\subsection{Implementation Details}
\label{subsec:Implementation Details}

In this work, we adopt Qwen2.5-VL-7B and Qwen2.5-VL-32B~\cite{bai2025qwen2.5vl} as our base models. 
For training data, we randomly sample 52K multimodal data from MM-Eureka~\cite{meng2025mm}. Model optimization is carried out using EasyR1~\cite{zheng2025easyr1} codebase, with training conducted on 8 NVIDIA H100 GPUs for the 7B model and 32 H100 GPUs for the 32B model.
For the rollout parameter, we use a question variant $m$ of 2, a sample number $n$ of 6 per question, and a probability $p$ of 0.3. For RL–related hyperparameters, we use a global batch size of 128, a rollout batch size of 512, a rollout temperature of 0.7, and a learning rate of 1e-6.

\subsection{Main Results}
\label{sec:exp_result}

To comprehensively examine the effectiveness of our proposed Share-GRPO, we conduct experiments on models of different sizes (\ie, 7B and 32B).
Notably, unlike prior studies~\cite{r1-vl,yang2025r1,huang2025vision}, we do not involve an additional cold-start stage with supervised fine-tuning.
As shown in Table~\ref{tab:main_result}, we provide an extensive comparison against state-of-the-art models across 6 widely used and challenging benchmarks, covering a diverse range of reasoning tasks from specialized domains to general-purpose reasoning. A detailed description of the benchmarks can be found in the appendix.

\input{table/main_table}

\textbf{Comparison with baselines.} We first compare our R1-ShareVL 7B and R1-ShareVL 32B trained by Share-GRPO with the corresponding base models, \ie, Qwen2.5-VL-7B and Qwen2.5-VL-32B. 
As presented in Table~\ref{tab:main_result}, 
Share-GRPO effectively improves the long-chain reasoning capabilities of MLLMs by large margins.
For example, on the challenging mathematical benchmarks like MathVista and MathVerse, R1-ShareVL-7B achieves improvements of +7.2\% and +3.6\%, respectively.
It is worth noting that, based on previous studies, RL can enhance MLLMs' long-chain reasoning ability on mathematical tasks, but it often comes at the cost of degraded performance on multi-discipline and general benchmarks. 
For instance, ThinkLite-7B drops -0.2\% and -5.5\% on MMStar and MMMU, respectively.
In contrast, our R1-ShareVL-7B model achieves a +3.1\% improvement on MMStar and comparable accuracy on MMMU, demonstrating Share-GRPO’s generalization capability in enhancing reasoning across diverse tasks.
When scaling our method to larger models (\ie, Qwen2.5-VL-32B) with stronger foundational capabilities, our method remains robust and consistently improves performance. In particular, R1-ShareVL-32B achieves a +9.1\% improvement over the baseline model on MathVerse, along with an average performance gain of +2.7\%.

\textbf{Comparison with MLLMs trained via RL.} 
We then compare R1-ShareVL with other state-of-the-art MLLMs trained by reinforcement learning approaches.
Our R1-ShareVL-7B using the same base model and fewer training data outperforms MM-Eureka-7B with an average performance gain of +2.1\%, especially a notable improvement of +1.4\% on MathVista.
Notably, beyond its capability in long-chain mathematical reasoning, R1-ShareVL also exhibits stronger reasoning generalization to multi-discipline and general reasoning tasks.
Specifically, compared to ThinkLite-7B which also excels in mathematical reasoning, R1-ShareVL achieves better performance on the multi-discipline benchmark MMMU and the general benchmark MMStar, outperforming it by +5.0\% and +3.3\%, respectively.
Besides, a similar conclusion can be observed on larger models: our R1-ShareVL 32B further improves overall performance compared with MM-Eureka-32B by +3.4\%, demonstrating the effectiveness and generalization of Share-GRPO.

\subsection{Ablation Study}
\label{sec:exp_ablation}

\begin{wraptable}{r}{0.6\textwidth}
\vspace{-2em}
\centering
\caption{\textbf{Ablation study of Share-GRPO.} }
\label{tab:noise_annealing_ablation}
\scalebox{0.65}{
\setlength{\tabcolsep}{4.0pt}
\begin{tabular}{lccccccccccccc}
\toprule
\multirow{2}{*}{Method} & \multicolumn{2}{c}{Shared Policy} & \multicolumn{2}{c}{Shared Advantage} & \multirow{2}{*}{MathVista} \\
    \cmidrule(lr){2-3} \cmidrule(lr){4-5}
    & \multirow{1}{*}{Offline} & \multirow{1}{*}{Online} & \multirow{1}{*}{Global} & \multirow{1}{*}{Local} &  \\
    \midrule
    Qwen2.5-VL-7B (Baseline) & & & & & 68.2 \\
    Qwen2.5-VL-7B + GRPO & & & & & 72.8 \\
    \midrule
    \multirow{3}{*}{Share-GRPO (Ours)} & \CheckmarkBold & & \CheckmarkBold & & 73.9 \\
     & \CheckmarkBold & \CheckmarkBold & \CheckmarkBold &  & 74.8 \\
     & \CheckmarkBold & \CheckmarkBold & \CheckmarkBold & \CheckmarkBold & 75.4 \\
\bottomrule
\vspace{-2em}
\end{tabular}
}
\label{tab: ablation study}
\end{wraptable}

\textbf{Ablation Study of Share-GRPO.}
As shown in Table~\ref{tab: ablation study}, we conduct ablation studies to examine the individual contribution of each design in Share-GRPO, including shared policy optimization (\ie, offline and online semantically consistent transformation) and shared advantage estimation (\ie, global and local advantage estimation).
Compared to the GRPO baseline, incorporating the information sharing among only offline question variants with global shared advantage estimation yields a performance boost of +1.1\%.
Further including the information sharing among online multimodal semantically consistent transformations results in exploring and sharing more diverse reasoning paths and a +0.9\% performance improvement.
Finally, enabling both global and local advantage estimation achieves the best result of 75.4\% on MathVista, highlighting the effectiveness of hierarchical advantage computation.
These results demonstrate that both policy sharing and advantage sharing contribute significantly to the final performance of Share-GRPO.

\subsection{Discussion}
\label{sec:exp_discussion}

\textbf{Complementarity Between Share-GRPO and Dynamic Sampling.} 
We compare Share-GRPO with dynamic sampling~\cite{yu2025dapo} and further discuss their complementarity, as shown in Table~\ref{tab: filter and share grpo}.
Dynamic sampling performs multiple rollouts to generate candidate responses, from which a subset with valid rewards is selected to update the policy.
Distinct from it, Share-GRPO enhances the diversity of reasoning responses and shares them during policy optimization, yielding richer reward signals.
On average, GRPO with dynamic sampling achieves a performance of 65.1\%, while Share-GRPO yields a higher score of 66.8\%, demonstrating the potential of leveraging diverse reasoning paths to enrich reward feedback and enhance policy learning.
Moreover, combining Share-GRPO with dynamic sampling achieves a higher accuracy of 67.2\% on average, highlighting the great complementarity of Share-GRPO.

\begin{table}[h]
\vspace{-1em}
\centering
\caption{The complementarity between Share-GRPO and Dynamic Sampling.}
\vskip 0.1in
\label{tab: filter and share grpo}
\scalebox{0.8}{
\setlength{\tabcolsep}{5pt}
\begin{tabular}{lccccccccccccc}
\toprule
    Method & MathVista & MMStar & MMMU & Avg. \\
    \midrule
    GRPO + Dynamic Sampling & 73.6 & 65.7 & 56.1 & 65.1 \\
    Share-GRPO & 75.4 & \textbf{67.0} & 58.1 & 66.8 \\
    Share-GRPO + Dynamic Sampling & \textbf{75.9} & 66.7 & \textbf{59.0} & \textbf{67.2} \\
\bottomrule
\end{tabular}
}
\end{table}

\textbf{Hyperparameter studies of Share GRPO.} 
We study the impact of the number of question variant $m$ in Semantically Consistent Transformation (SCT) in Table~\ref{tab:Ablation Study on question number.}, with the number of responses per question fixed at 6.
When $m=2$, Share-GRPO achieves an accuracy of 75.4\% on MathVista, significantly outperforming the baseline GRPO.
When $m$ is increased to $4$, the performance of the model trained with Share-GRPO further improves, demonstrating the benefit of scaling up the number of question variants.

\begin{figure*}[t]
  \centering
  \includegraphics[width=\textwidth]{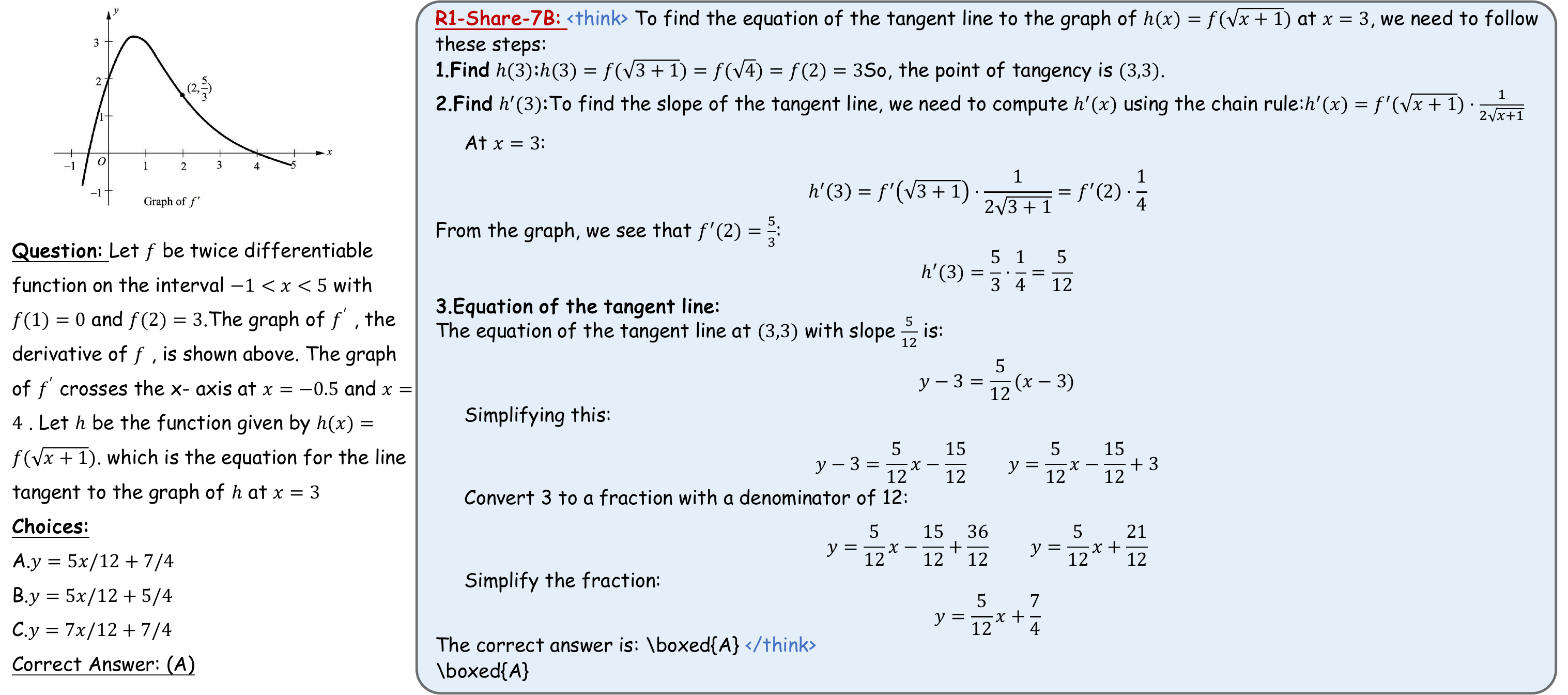}  
  \caption{\textbf{Qualitative Results} of reasoning capability of R1-ShareVL on mathematical problems.}
  \label{fig:qualitative}
  \vskip -0.1in
\end{figure*}

\begin{table}[t]
\centering
\begin{minipage}[t]{0.43\textwidth}
  \centering
  \caption{\textbf{Hyperparameter study of Share-GRPO.} We study the impact of the number of question variants in Share-GRPO.}
  \vskip 0.1in
  \scalebox{0.9}{
  \setlength{\tabcolsep}{6.pt}
  \begin{tabular}{lccccccccccccc}
\toprule
    Method & Question & MathVista \\
    \midrule
    ShareGRPO & 2 & 75.4 \\
    ShareGRPO & 3 & 75.6 \\
    ShareGRPO & 4 & 75.9 \\
\bottomrule
\end{tabular}}
  \vskip -0.1in
  \label{tab:Ablation Study on question number.}
\end{minipage}%
\hspace{0.1\textwidth}
\begin{minipage}[t]{0.43\textwidth}
  \centering
  \caption{\textbf{The study on the Number of Sampling N}.  We study the impact of the number of sampling N in Share-GRPO.}
  \vskip 0.1in
  \scalebox{0.85}{
  \setlength{\tabcolsep}{5.pt}
  \begin{tabular}{lccccccccccccc}
\toprule
    Method & Sampling & MathVista \\
    \midrule 
    GRPO & 6 & 72.3 \\
    GRPO & 12 & 72.8 \\
    GRPO & 24 & 73.0 \\
    \midrule
    ShareGRPO & (3+3) & 74.7 \\
    ShareGRPO & (6+6) & 75.4 \\
\bottomrule
\end{tabular}}
  \vskip -0.2in
  \label{tab: Ablation Study on sampling number.}
\end{minipage}
\end{table}

\textbf{The Impact of the Number of Sampling N.} 
In Table~\ref{tab: Ablation Study on sampling number.}, we compare the performance of GRPO with different sampling numbers $N$ against Share-GRPO. 
We observe that with $N=6$, the model trained using GRPO achieves a score of 72.3\% on MathVista.
As the number of sampling increases, the performance improves to 72.8 at $N=12$.
However, further increasing the sampling number $N$ to 24 yields only marginal gains of 0.2\%, while introducing additional computational overhead.
Therefore, increasing the number of sampling reaches a performance ceiling, making it an ineffective way to further improve reasoning reinforcement learning.
Instead of simply increasing $N$, Share-GRPO enhances the diversity of reasoning paths and leverages the concept of information sharing to amplify reward signals and enhance training stability.
By sharing responses and incorporating hierarchical advantage estimation, our R1-ShareVL 7B achieves a score of 75.4\% with only 6 generated responses per question, surpassing the performance of GRPO even with 24 sampled responses.

\subsection{Qualitative Results} 
Fig.~\ref{fig:qualitative} illustrates that Share-GRPO effectively enhances the model’s reasoning ability on complex mathematical problems. In this example, the model accurately interprets the question and arrives at the correct answer, showing strong performance in symbolic reasoning and function analysis.
This highlights the capability of Share-GRPO to guide the model toward precise and coherent solutions in mathematically demanding tasks.

\section{Conclusion}
In this paper, we propose Share-GRPO, a novel reinforcement learning framework for MLLMs, which introduces the concept of information sharing to effectively mitigate the challenges of sparse rewards and advantage vanishing.
Share-GRPO expands the question space by generating semantically consistent variants, and encourages MLLMs to explore and share responses across a more diverse solution space.
Furthermore, Share-GRPO estimates advantages hierarchically within and across question variants at both global and local levels to effectively guide optimization.
We conduct extensive experiments, ablation
studies and discussion, which demonstrate the superiority of our proposed methods on various reasoning benchmarks.

\clearpage

{
\small
\bibliographystyle{unsrt}
\bibliography{reference}
}

\medskip

\clearpage
\appendix

\section{Benchmarks}
We evaluate our models on the following benchmarks.
\begin{itemize}
    \item \textbf{MathVista~\cite{mathvista}} is used to evaluate the mathematical problem-solving ability of MLLMs, containing 6141 questions covering areas such as arithmetic, geometry, algebra, and statistics.
    \item \textbf{MMStar~\cite{mmstar}} is an innovative multimodal assessment benchmark that includes 1500 carefully selected visual key samples, addressing issues of visual redundancy and data leakage in existing assessments.
    \item \textbf{MMMU~\cite{mmmu}} is a large-scale interdisciplinary multimodal understanding and reasoning benchmark that collects 11.5K multimodal questions from university exams, quizzes, and textbooks.
    \item \textbf{MathVerse~\cite{zhang2024mathverse}} includes 2612 multimodal mathematics problems and has manually annotated 15672 test samples, comprising 3 main types of questions and 12 subcategories, such as plane geometry, solid geometry, and functions.
    \item \textbf{MathVision~\cite{mathvision}} is a collection of 3,040 high-quality mathematics problems, all accompanied by visual contexts, sourced from real mathematics competitions.
    \item \textbf{AI2D~\cite{kembhavi2016diagram}} is a dataset that contains over 5000 scientific charts, which can be used for tasks such as image classification and visual question answering.
\end{itemize}

\end{document}

%% file: table/main_table.tex
\begin{table}[t]
    \caption{\textbf{Main Results.} To examine the effectiveness of Share-GRPO, we compare our R1-ShareVL which is trained by Share-GRPO without cold-start supervised fine-tuning against SOTAs across multiple reasoning tasks, including both domain-specific and general-purpose tasks. $^*$ denotes evaluation on official weights using VLMEvalKit~\cite{duan2024vlmevalkit}.}
    \vskip 0.1in
    \centering
    \scalebox{0.85}{
    \setlength{\tabcolsep}{3pt}
        \begin{tabular}{lccccccccc}
            \toprule
            \textbf{Model}  & \textbf{MathVista} & \textbf{MMStar} & \textbf{MMMU} & \textbf{MathVerse} & \textbf{MathVision} & \textbf{AI2D} & \textbf{Avg.} \\
            \midrule
            GPT-4o\cite{gpt-4o} & 63.8 & 65.1 & 70.7 & 50.8 & 30.4 & 84.9 & 60.9\\
            Claude3.7-Sonnet\cite{claude_3.5_sonnet} &  66.8 & -- & 71.8  & 52.0 & 41.3 & -- & -- \\
            Kimi1.5\cite{team2025kimi} &  70.1 & -- & 68.0 & -- & 31.0 & -- & -- \\
            \midrule
            LLaVA-Reasoner-8B~\cite{llava-rasoner} & 50.6 & 54.0 & 40.0 & -- & -- & 78.5 & -- \\
            LLaVA-CoT-11B\cite{xu2411llava} & 54.8 &  57.6 & -- & --& --& 78.7 & -- \\
            Mulberry-7B\cite{yao2024mulberry}&   63.1 & 61.3 & 55.0 & -- & -- & -- & -- \\
            \midrule
            Qwen2.5-VL-7B \cite{qwen2_5_vl} (Base Model)  & 68.2 & 63.9 & 58.6 & 49.2 & 25.1 & 83.9 & 58.1 \\
            X-REASONER-7B \cite{xreasoner} & 69.0 & -- & 56.4 & -- & 29.6 & -- & --\\ 
            R1-Onevision-7B\cite{yang2025r1} & 64.1 & -- & -- & 47.1 & 29.9 & -- & -- \\  
            Vision-R1-7B\cite{huang2025vision}  &  73.5 & ~~64.3$^*$ & ~~54.2$^*$ & 52.4 & ~~29.4$^*$ &~~84.2$^*$ & 59.7\\ 
            OpenVLThinker-7B\cite{deng2025openvlthinker} & 70.2 & 63.2 & 51.9 & 47.9 & 29.6 & 82.7 & 57.6 \\
            MM-Eureka-7B\cite{meng2025mm} &  73.0 & ~~65.1$^*$  & ~~55.3$^*$ & 50.3  & 26.9 & ~~84.1$^*$ 
            & 59.1 \\
            ThinkLite-7B \cite{thinklite-vl}& 74.3 & 63.7 & 53.1 & 52.2 & 29.9 & 83.0 & 59.3 \\
            \midrule
            \rowcolor{mygray} R1-ShareVL-7B &  75.4 & 67.0 & 58.1 & 52.8 & 29.5 & 84.5 & 61.2 \\
            \midrule
            \midrule
            \multicolumn{9}{c}{\small \em Scaling to Larger Models }\\ \midrule
            
            Qwen2.5-VL-32B \cite{qwen2_5_vl} (Base Model) & 74.7 & 69.5 & 70.0 & 49.9 & 38.4 & ~~84.6$^*$ & 64.5 \\
            MM-Eureka-32B\cite{meng2025mm}& 74.8 & ~~67.3$^*$  & ~~64.6$^*$ & 56.5  & 34.4 &  ~~85.4$^*$ & 63.8 \\
            \midrule
            \rowcolor{mygray} R1-ShareVL-32B & 77.6 & 70.2 & 70.1 & 59.0 & 40.3 & 86.2 & 67.2 \\
            \bottomrule
        \end{tabular}
        }
    \label{tab:main_result}
    \vskip -0.2in
\end{table}